# Theoretical foundations and limits of word embeddings: what types of meaning can they capture?

Running head: Theoretical foundations and limits of word embeddings

Alina Arseniev-Koehler[*]

Department of Sociology, University of California, Los Angeles

---

[*] Please direct all correspondences to Alina Arseniev-Koehler (arsena@g.ucla.edu). Address: Haines Hall, 264, 375 Portola Plaza, Los Angeles, CA 90095. This paper is based on work supported by the National Science Foundation Graduate Research Fellowship Program under Grant No. DGE-1650604. I am grateful to Omar Lizardo, Jacob Foster, Bernard Koch, Eleni Skaperdas, Sam Ashelman, Devin Cornell, the members of the UCLA Conversation Analysis Working Group, and anonymous reviewers for their feedback on this work.

Theoretical foundations and limits of word embeddings: what types of meaning can they capture?

Measuring meaning is a central problem in cultural sociology and word embeddings may offer powerful new tools to do so. But like any tool, they build on and exert theoretical assumptions. In this paper I theorize the ways in which word embeddings model three core premises of a structural linguistic theory of meaning: that meaning is relational, coherent, and may be analyzed as a static system. In certain ways, word embedding methods are vulnerable to the same, enduring critiques of these premises. In other ways, they offer novel solutions to these critiques. More broadly, formalizing the study of meaning with word embeddings offers theoretical opportunities to clarify core concepts and debates in cultural sociology, such as the coherence of meaning. Just as network analysis specified the once vague notion of social relations (Borgatti et al. 2009), formalizing meaning with embedding methods can push us to specify and reimagine meaning itself.

Measuring, modeling, and understanding how meaning operates are several of the most prominent and longstanding endeavors of sociology (e.g., Mohr 1998; Mohr et al. 2020). In recent years, word embedding methods reinvigorated the study of meaning (e.g., Arseniev-Koehler and Foster 2020; Boutyline, Arseniev-Koehler, and Cornell 2020; Charlesworth et al. 2021; Garg et al. 2018; W. Hamilton, Leskovec, and Jurafsky 2016; Jones et al. 2020; Kozlowski, Taddy, and Evans 2019; Nelson 2021; Stoltz and Taylor 2019, 2020; Taylor and Stoltz 2020). These methods computationally model the semantic information of words in large-scale text data. Despite their promise, it remains unclear what kind of meaning word embeddings capture – or whether they capture any meaning at all. If we are to employ these tools rigorously, it is paramount that we clarify what they operationalize. In this paper, I critically evaluate the possibility that word embeddings operationalize an influential theory of linguistic meaning: linguistic structuralism.

Word embeddings, and in particular the word2vec word embedding algorithm, revolutionized how computers learn and process human language (Mikolov, Chen, et al. 2013; Mikolov, Sutskever, et al. 2013). Indeed, since word2vec was published in 2013 it has been cited well over 28,000 times (Mikolov, Sutskever, et al. 2013). Now, word embedding is becoming a pervasive text analysis approach in social science (for a review, see Stoltz and Taylor 2021). For example, these methods are used to capture stereotypes encoded in media language across time, offering a historical lens stereotypes despite the absence of corresponding survey data. This paper focuses on word embeddings given their popularity in social science; however, their premises may be generalized beyond language to model other cultural systems (e.g., Arronte Alvarez and Gómez-Martin 2019; Chuan, Agres, and Herremans 2020). In sum, word embeddings offer exciting new lenses into language – and perhaps other cultural systems – across time and space.

Researchers employing word embedding methods frequently note their affinities to a century-old theoretical perspective on language: linguistic structuralism (e.g., Baroni, Dinu, and Kruszewsk 2014; Faruqui et al. 2016; Günther, Rinaldi, and Marelli 2019; Kozlowski et al. 2019). Linguistic structuralism envisions language as a symbolic system comprised of various linguistic units (e.g., words or suffixes). These units' meanings are defined by their relationships to other units in the system (rather than by their reference to physical objects in the world). For instance, a word's meaning is defined by its relationships to other words in the vocabulary. Scholars ultimately generalized linguistic structuralism to study non-linguistic symbolic systems– such as roles in a kinship system (Lévi-Strauss 1963:35). This intellectual movement is often referred to as semiotic structuralism, French structuralism, or just "structuralism" more broadly. Here, I critically examine the affinities between the way that word embeddings model words' semantic information and a linguistic structuralist perspective on word meaning. I focus on how word embeddings do and do not operationalize each of several core premises of linguistic structuralism. The first major contribution of this paper is to highlight that the extent to which contemporary word embedding methods operationalize linguistic structuralism depends on the way these methods are used, the specific embedding algorithm used, and even an analysts' own interpretation of "meaning" in the algorithm.

Linguistic structuralism (and structuralism more broadly) was both profoundly influential and intensely critiqued (Dosse 1997). For instance, critics questioned the extent to which a shared, coherent symbolic system exists (e.g., Martin 2010; Swidler 1986), noting that cultural symbols (like words) are often used in strikingly incoherent ways (e.g., Swidler 2013). Given the parallels between word embeddings and structuralism, do word embeddings also model language in a way that is vulnerable to the critiques of structuralism? The second major contribution of this paper is to evaluate the ways in which word embeddings *succumb to* or *overcome* the limitations of linguistic structuralism.

To begin, I first briefly review background information on word embedding methods in social science. Second, I review linguistic structuralism, focusing on three of its core premises (that language is relational, coherent, and can be analyzed as a static system). Third, I critically examine the ways in which word embeddings operationalize each of these three premises. Fourth, I consider four critiques of these premises and evaluate the extent to which each critique applies to word embeddings. In the discussion, I highlight implications and directions for future sociological research with word embedding methods.

## 1. A PRIMER ON WORD EMBEDDINGS

The term "word embedding" refers to numerous approaches to quantitatively represent the semantic information of words based on how those words are used in a text dataset (e.g., a corpus of news articles, social media posts, government records, or product reviews). Contemporary word embeddings aim to represent words as vectors (i.e., arrays of *N* numbers) where words that are used in more similar contexts in the data tend to be assigned more similar vectors. Word vectors may also be understood graphically, as positions in space. Just as a 2-dimensional vector locates a position in 2-dimensional space (where this space is a Euclidean plane, defined by an "X" and a "Y" axis), an *N*-dimensional vector locates a position in an *N*-dimensional space.[1] Thus, each word vector may be thought of as locating a word in *N*-dimensional semantic space. The dimensionality of the space (and thus all word vectors) is pre-set by the algorithm or the

researcher – often, at a few hundred dimensions (Rong, 2014). The information captured by dimensions are latent; they are identified by word embedding algorithm as organizes words in space. Since vectors locate positions in space, similarity and distance are interchangeable. Words with more similar vector representations are also *closer* in space. This similarity, or distance, is commonly measured with cosine similarity.

A variety of results indicate that word embeddings can accurately capture semantic information of words, based on the way that words are used in natural language. For example, in well-trained word embeddings, the cosine similarities between word vectors tightly correlate to human-rated similarities between words (e.g., Caliskan, Bryson, and Narayanan 2017; Pennington, Socher, and Manning 2014). Further, while word embeddings are trained to represent words as positions in space (i.e., as vectors), empirical work finds that the space *between* word vectors may capture semantic information.

Most famously, the direction to travel in semantic space between the word vectors "king" and "queen" is often similar to the direction to travel between "man" and "woman." That is, the difference between the locations of "king" and "queen" in semantic space is similar to the difference between "man" and "woman" (Mikolov, Chen, et al. 2013; Mikolov, Sutskever, et al. 2013). The difference may be measured by subtracting the corresponding word vectors, e.g., "woman" – "man." The result of this subtraction is a vector that may be interpreted as a latent line in space, pointing to femininity at one pole and masculinity at the other pole (see figure 1). In fact, a variety of concepts beyond gender may be encoded as latent dimensions in space (Bolukbasi et al. 2016; Kozlowski et al. 2019; Mikolov, Sutskever, et al. 2013). This property of word embeddings attests to their ability to encode semantic information in rich and nuanced ways. Furthermore, as described next, this property makes word embedding a useful social science method.

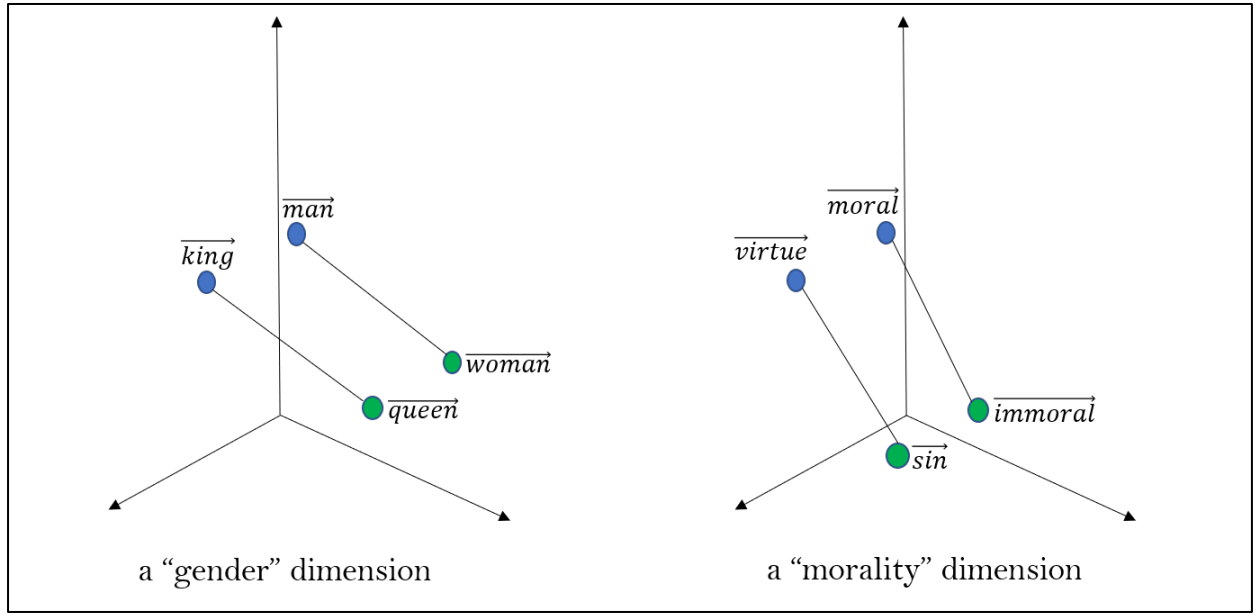

*Figure 1.* Conceptual illustration of a latent dimension in semantic space corresponding to gender (left) and morality (right).

### 1.1. Word Embeddings as Social Science Methods

In recent years, word embedding has exploded as an exciting new method in social science (e.g., Arseniev-Koehler and Foster 2020; Boutyline et al. 2020; Charlesworth et al. 2021; Garg et al. 2018; W. Hamilton et al. 2016; Jones et al. 2020; Kozlowski et al. 2019; Nelson 2021, 2021; Stoltz and Taylor 2019, 2020; Taylor and Stoltz 2020). A predominant analytic approach to identify a latent dimension (e.g., gender, social class, or morality) in semantic space, and then examine how a set of sociologically interesting words are positioned along this dimension. For example, researchers may extract a line corresponding to gender, by subtracting the word vector "she" from "he," and then examine how close occupational terms are to the feminine or masculine pole of this line (Bolukbasi et al. 2016). Effectively, this approach enables analysts to compute the association of a word (or set of words) with a latent concept in text data, such as the extent to which a word is described as feminine or masculine (e.g., Arseniev-Koehler and Foster 2020; Kozlowski et al. 2019).

As one example of a sociological application of word embeddings, Jones et al. used this approach to investigate the gendered associations of words about career, family, science, and art, using word embeddings trained across two centuries of books (2020). Their findings suggest that that many of the gender stereotypes in these domains have receded in language over time. As a second example, Kozlowski et al. used word embeddings to investigate five dimensions of meanings relevant to social class in books across the twentieth century (2019). Their results suggested that the cultural associations of education with social class emerged only in recent decades, while in the earlier part of the 20$^{th}$ century these associations are mediated by meanings of cultivation. As these studies illustrate, word embeddings enable social scientists to investigate cultural phenomena in ways that may be impractical or even impossible using surveys or other traditional social science methods.

Social scientists have also begun to use word embedding methods to investigate the relationships between language, widely held personal meanings (e.g., from survey responses), and broader societal patterns (e.g., demographic changes). For instance, Caliskan et al. illustrated the correspondences between various implicit associations in human participants[2] and cosine similarities between corresponding word vectors (2017). Further, Garg et al. showed that the way occupations are gendered in word embeddings corresponds tightly (correlations around

*DRAFT*                                                     *Theoretical Foundations and Limits of Word Embeddings*.9) to the proportion of women in these occupations based on Census data, both in present day and across time (2018). Both papers identified these patterns across embeddings trained on a variety of corpora. As these and other examples illustrate (e.g., Boutyline et al. 2020; Jones et al. 2020; Kozlowski et al. 2019), word embeddings also offer a new approach investigate links between cultural, social, and linguistic change.

Despite the promise and predominance of word embeddings, social scientists are just beginning to reconcile these methods with longstanding theoretical work on meaning and culture (e.g., Arseniev-Koehler and Foster 2020; Kozlowski et al. 2019). To theorize these methods, it is important to first understand how we arrive at a trained word embedding from raw text data. These methodological details are reviewed next.

*1.2. Approaches to Compute Word Embeddings Commonly used in Social Science*

A wide variety of algorithms are used to compute word embeddings in social science. These algorithmic differences are important because they have implications for theorizing the kind of meaning that word embeddings operationalize. Most crucially, word embeddings are computed using count-based approaches or using a machine-learning framework called artificial neural networks (i.e., neural word embeddings) (Baroni et al. 2014). I briefly describe each next.

Count-based approaches begin with a word by word (or word by document) co-occurrence matrix computed from the corpus and attempt to reduce the dimensionality of this matrix by finding *N* latent, lower-dimensional features that encode most of the matrix's structure. A wide variety of methods have been used for dimensionality reduction. The output from performing dimensionality reduction is a word by *N* dimensional matrix: each row is an *N*-dimensional word vector. Among the most popular and successful count-based word embedding approaches is GloVe (Pennington et al. 2014). However, since the publication of word2vec (Mikolov, Chen, et al. 2013; Mikolov, Sutskever, et al. 2013), neural word embedding architectures are becoming dominant in computer science for their flexibility and performance on downstream tasks (Baroni et al. 2014).[3]

Neural word embeddings use artificial neural networks to *incrementally learn* word embeddings from a given corpus as they "read" the text data and attempt to predict missing words in the data. For example, in word2vec[4] with Continuous Bag-of-Words (word2vec-CBOW), the model learns word vectors as it attempts to "fill-in" missing words from various sets of contexts in the text (Mikolov, Chen, et al. 2013; Mikolov, Sutskever, et al. 2013). Word2vec-CBOW is iteratively given a context of words (e.g., 10 words) with one word missing, and is tasked with predicting the missing word. To make a prediction, the model first averages the observed context word vectors to form a single vector representing the context.[5] Second, it predicts the missing word based on the word vector which is *most similar* (or closest in space) to this context vector.[6] Since word vectors may be initially randomized, the model initially tends to predict the missing words incorrectly. Each time it guesses incorrectly, the correct word is revealed, and the model updates the word vectors to reduce this prediction error (and thus, improve its chances at guessing correctly if it were to see this context again). As word2vec adjusts word vectors across thousands of attempts to predict words from their contexts in the data, the word vectors begin to better represent words. Upon reaching some pre-determined stopping point (e.g., the level of accuracy), we can stop training and use the most recent word vectors for downstream analyses. Thus the result (like count-based embeddings) is a word by *N* dimensional matrix: each row is an *N*-dimensional word vector.

Contemporary word embeddings were developed to enable computers to learn, process, and represent human language, not to operationalize any theory of language or linguistic meaning. As these methods gain traction in social science it is crucial that we clarify what they do and do not operationalize. This paper values the possibility is that word embeddings operationalize an influential but controversial theoretical perspective on words' meanings: linguistic structuralism.

## 2. A PRIMER ON LINGUISITIC STRUCTRURALISM

Linguistic structuralism is a theoretical perspective on language, and, more broadly, an approach to study language (Craib 1992:131–48; Joas and Knobl 2009:339–70). In this paper, I focus on three of its key premises. First, that language is a system comprised of various signs (e.g., words, suffixes, or idioms) where these signs are purely defined by their *relationship* to other signs in the system, rather than by any external reality. For example, a word is defined by its co-occurrence relationship to all other words – *not* from the intrinsic properties of the letters or sounds that comprise the word, from dictionary definitions, or by its reference to some external object (Saussure, 1983, p. 113). This suggests, for example, that if a misspelled word is *used* in a similar way as a correctly spelled word, both spelling variants will be understood in the same way. If, however, spelling variants are used in some systematically different way (e.g., British versus American spellings), the variants will evoke slightly different interpretations– even when all variants refer to the same physical object.

Identifying and understanding the relationships in this system is the core goal of linguistic structuralism. One well-theorized type of relationship is a binary opposition (Lévi-Strauss, 1963, p. 35): a structure of meaning where two concepts are defined by their oppositional relationship to one another. In this perspective, for example, we cannot conceive of the concept of "good" without that of "evil" because they form a binary opposition. "Good" is defined by its difference from "evil" and vice versa. Theoretically, structuralism suggests that binary opposition is a key, latent structure of meaning which scaffold symbolic systems, such as language. Therefore, a common empirical goal in linguistic structuralism (and structuralism more broadly) is to identify binary oppositions in a symbolic system (e.g., Barthes 2012; Lévi-Strauss 1963:35).

A second core premise of linguistic structuralism is that underlying the inconsistent ways in which we use words, there exists a latent, coherent linguistic system.[7] This system is psychological (i.e., internalized), but also generalized beyond any individual language user and thus shared, i.e., cultural (see also Stoltz 2019). Linguistic structuralism focuses on studying this hypothesized cultural system rather than explaining the varying ways in which we use language.

A third core premise of linguistic structuralism is the distinction between studying language as a static versus dynamic system.[8] While the first considers how the parts within a linguistic system interact at any given point (e.g., what are the kinds of the relationships that exist between words), the second focuses on how this system changes and why (e.g., how words' positions in the system change or how new words emerge). Analogously, one can study chess as a static or dynamic system: we can "freeze" a chess game and describe where the pieces lie on the chess board in relation to one another, or we can describe the movements of pieces across a game. Linguistic structuralism focuses theoretical and analytical efforts on language as if static, studying language as a "snapshot" in time.

## 3. OPERATIONALIZING LINGUISTIC STRUCTURALISM WITH WORD EMBEDDINGS

Here, I detail the ways in which word embeddings may be used to operationalize each of the three core premises of linguistic structuralism described previously: the focus on language as a relational, coherent, and static system. Here, I consider these three premises as they pertain to the meanings of words. However, linguistic structuralism applies to various aspects of language (e.g., the sounds of letters and the meanings of words) and was ultimately generalized to study non-linguistic symbolic systems (for a more detailed review, see Craib 1992:131–48).

*3.1. Modeling Language as a Relational System with Word Embeddings*

Word embeddings methods can operationalize a relational notion of linguistic meaning several ways. First, word embeddings *learn* word vectors relationally. Specifically, they rely on a hypothesis in linguistic structuralism that words are used in more similar ways in a corpus will share similar meanings (Firth 1957; Harris 1954). This is called the Distributional hypothesis and is more colloquially described as "a word is characterized by the company it keeps." This hypothesis is also fundamentally relational, suggesting that a word may be understood by its co-occurrence relationships with other words. All approaches to train word embeddings operationalize the Distributional hypothesis in some form or another. For instance, word2vec-CBOW predicts words from their various contexts in the corpus, and moves words to be closer to (i.e., more similar to) their context vectors. GloVe begins with a matrix containing the co-occurrences between words in a corpus and compresses this matrix to arrive at word vectors. Thus, these models are relational in that they learn word vectors as defined by relationship between each word is used in the corpus and how all other words are used in the corpus.

However, the extent to which word embeddings operationalize a relational theory of meaning depends on the analyst's interpretation of "meaning" in the Distributional Hypothesis. Indeed, the Distributional hypothesis is notoriously vague when it comes to "meaning" (Sahlgren 2008); "meaning" is only relevant when this concept is introduced by the analyst. In practice, researchers often implicitly interpret the Distributional hypothesis somewhere along two extremes (Sahlgren 2008). At one extreme lies the weakest reading of the hypothesis: a word's meaning – whatever that might be – *correlates* to the patterned ways in which it is used in language. In this reading, "meaning" is latent: word embeddings are not necessarily capturing meaning, let alone a relational notion of meaning. At the other extreme, in a stronger reading of the hypothesis, the meaning of a word *is* the patterned ways in which it is used natural language (rather than, say, its dictionary definition, or reference to a physical object). This second interpretation is more structuralist because it suggests that words are defined relationally, rather than by any referent external to the linguistic system. This second interpretation is also cognitive and causal, suggesting that words' meanings are *learned* from their relational patterns in natural language. A midrange ("partly structuralist") interpretation might be that meaning is, in part, defined by the relational pattern of words.

Second, word embeddings not only learn words in a relational fashion, but also represent words relationally. As described in section one, word embeddings represent words by their position along each of several hundred dimensions (i.e., by N-dimensional word vectors). Word vectors are only interpretable to analysts because of their *relative* positions, such as where "man" lies in relation to all other word vectors in the vocabulary. Because word vectors are not tied to any external referents and defined only within their self-contained semantic space, any given word vector is arbitrary and uninformative outside of this space. Like with the Distributional hypothesis, the *extent* to which word embeddings operationalize a relational theory of meaning varies based on the analyst's interpretation of "word vectors." Analysts may simply use word vectors as tools to represent meaning, without presuming that word vectors are accurate metaphors for words' meanings (e.g., Caliskan et al. 2017; Charlesworth et al. 2021; Garg et al. 2018; Jones et al. 2020). Or, analysts may also interpret word vectors as cognitively plausible representations of words meanings (e.g., Arseniev-Koehler and Foster 2020). This second interpretation is also more strongly structuralist.

Third, social scientific analyses using word embeddings often identify latent concepts in semantic space as binary oppositions, a key relationship posited by linguistic structuralism. For instance, as described in part one, a core approach to measure a concept, like gender, is to identify a line between the word vectors for two opposing poles (e.g., "woman" and "man" for gender). This measure operationalizes the concept of gender as ranging continuously from one pole to the other. Being closer to one pole implies being farther from the other (e.g., more masculinity implies less femininity). In a related approach, analysts identify word vectors corresponding to two poles of an opposition (e.g., "woman" and "man" to represent gender) and then compare the distance of some interesting word to each pole. While this second approach does not assume that more femininity implies less masculinity or that gender is represented as a line in semantic space, both approaches identify concepts as oriented by two poles and thus operationalize a binary opposition. Just as identifying and studying binary oppositions (e.g., gender, morality, and sentiment) is one core focus of linguistic structuralism, it is also a core approach in current social science work using embeddings (e.g., Arseniev-Koehler and Foster 2020; Boutyline et al. 2020; Caliskan et al. 2017; Garg et al. 2018; Jones et al. 2020; Kozlowski et al. 2019; Nelson 2021; Taylor and Stoltz 2021).

*3.2. Modeling Language as a Coherent System with Word Embeddings*

Following a linguistic structuralist perspective, word embeddings abstract a latent semantic system from the diverse ways in which words are used in a training corpus. Word embeddings are geared towards modeling language as coherent in two specific ways. First, word embedding methods commonly used in social science represent each word as a single word vector. Thus, these word vectors may be thought of as capturing the regularities over all the various contexts in which the word appears. For example, the word vector for "bank" captures patterns across the diverse ways "bank" may be used across training data, across the different writers who generated the documents, and across geographic regions where these documents were generated.

Second, the architecture of word embeddings presumes that there are regularities across vocabulary words, and so limited number of dimensions can represent all words in a vocabulary. More specifically, words are represented as vectors where each element corresponds to a loading on each of $N$ dimensions, as described in section one. The dimensionality ($N$) of word vectors is far lower than the vocabulary size itself ($V$). Often, $N$ is set at a few hundred, while vocabulary sizes often range from tens of thousands to several hundred thousand words depending on the corpus. This difference in sizes *assumes* that there are patterns across the co-occurrences of words which will accurately capture a high dimensional vocabulary. Dimensions are shared and reused across vocabulary words to represent different aspects of their meaning, and so only a limited number of dimensions is needed to model words. Thus, in a structuralist fashion, word embedding methods aim to extract a hypothesized, latent system across the diversity of words and ways these words are used, and analyses then focus on understanding this system.

*3.3. Modeling Language as a Static System with Word Embeddings*

Word embeddings may also be used to model language as a static system, following a structural linguistic perspective. Neural and count-based embeddings do so in different ways. As described in section one, count-based embeddings abstract a latent semantic space from a co-occurrence matrix, using some dimensionality reduction algorithm. Neural word embeddings learn a semantic space as they attempt to predict missing words while "reading" data. When we *use* word vectors from neural embeddings in social science applications, we generally stop the training process (upon reaching some threshold, such as sufficient predictive accuracy)[9] and begin analyses on our "snapshot" of the system. Thus, all word embedding can be used to examine language as a static system, but count-based and neural word embeddings do so in different ways.

  Social scientific analyses using word embeddings vary in the extent to which they focus on language as static or dynamic. While some work investigates word embeddings at a single time point (e.g., Arseniev-Koehler and Foster 2020; Caliskan et al. 2017), a growing body of scholarship uses word embeddings track shared stereotypes and biases across time at macro-scales, such as across years or even decades (e.g., Boutyline et al. 2020; W. L. Hamilton, Leskovec, and Jurafsky 2016; Jones et al. 2020; Kozlowski et al. 2019). While embeddings offer exciting opportunities for both analyses across time and within a given time point, the latter might be considered a more distinctly structuralist approach.

## 4. FOUR CRITIQUES OF LINGUISTIC STRUCTURALISM

Here, I consider four critiques of linguistic structuralism: (1) that meaning may be grounded or embodied, rather than purely relational, (2) that a focus on binary oppositions is reductionistic, (3) that meaning is incoherent, and (4) that language is dynamic. After briefly introducing each critique, I argue how it applies (or does not apply) to word embeddings.

*4.1. Critiques of Purely Relational Approaches to Meaning*

While linguistic structuralism theorizes meaning as purely relational, scholars point to evidence it is linked to concrete referents in the external world: physical objects and events, sensorimotor experiences, and/or emotional experiences (e.g., Lakoff and Johnson 2008; Moseley et al. 2012; Pulvermüller 2013; Quiroga et al. 2005; Smith and Gasser 2005). For example, we can know when and how to use the word "recession" next to words like "money" and "finance." But we can also understand when "recession" refers to a specific event in history. We can then identify other, non-linguistic references to the same event, such as from viewing a chart of stock prices. This is possible partly because our brains incorporate semantic information from a variety of sources (Quiroga et al. 2005), not merely from language. As this critique highlights, it is unrealistic to theorize language as a self-contained system of signs (see also Bryson 2008).

  *4.1.1. Do these critiques apply to word embeddings?*

The extent to which critiques of a relational approach to meaning apply to word embeddings partly depends on the analysts' interpretation of "meaning" in word embeddings. As highlighted in section 3.1, word embeddings rely on the Distributional Hypothesis to learn words' semantic information, but analysts' interpretations of "meaning" in the Distributional hypothesis vary. A weaker reading of the Distributional hypothesis is that a word's meaning correlates to its relationship to other words in a language. This interpretation is not structuralist and so critiques of a relational vision of language do not apply. A stronger reading of the Distributional hypothesis is that a word's meaning *is* defined by its relationship to other words in a language. This stronger interpretation is structuralist, and vulnerable to broader critiques that linguistic meaning may also be grounded or embodied, rather than merely relational.

While critics point to the weakness of a purely relational notion of meaning, some words do not have concrete references or are not easily experienced. For example, not all humans have experienced the words "dive," "depression," or "royalty," and yet we know exactly how to use these words to transmit and build up larger ideas. Further, abstract words like "epistemic" and "subjective" are unlikely to be learned from sensorimotor experience or physical objects. Indeed, word embeddings (which learn from language alone) suggest a *mechanism* for learning and communicating such abstract concepts: the relational patterns of words in language (Borghi et al. 2017; Günther et al. 2019). Perhaps, a mid-range reading of the distributional hypothesis is warranted: that meaning correlates to distributional patterns of words and meaning may be also partly learned from distributional patterns in language (Lenci 2008).

In fact, a stream of work in computer science aims to develop language models where meaning is both relational (learned from text data, like in traditional word embeddings), and is learned from extra-linguistic experiences, such as images of what a word represents (Baroni 2016; Bruni, Tran, and Baroni 2014; Goh et al. 2021; Li and Gauthier 2017; Radford et al. 2021; Roy 2005; Vijayakumar, Vedantam, and Parikh 2017). Like humans, these multimodal word embeddings integrate semantic information derived from text, images, sound, or other modalities. They capture slightly different information than may be learned from text alone (Vijayakumar et al. 2017). For example, while word2vec learns the word "apple" as closest in space to "apples," "pear," "fruit," and "berry," a word2vec model also trained on sounds learns "apple" as closest to "bite," "snack," "chips," and "chew" (Vijayakumar et al. 2017). Multimodal word embeddings are not yet popular in social science but offer another way for word embeddings to break from critiques a purely relational approach to model words (see also Bisk et al. 2020).

*4.2. Critiques of the Structuralist Focus on Binary Oppositions*

The structuralist focus on binary oppositions has also received extensive critique. Binary oppositions are certainly not the only forms for meaning, and opposition itself come in many varieties: hierarchical, continuous, dichotomous, or graded (Geeraerts 2010:87). For instance, we might describe aesthetics as a dichotomous concept (as beautiful versus ugly), or on a gradable scale (ugly versus plain versus pleasant versus gorgeous). Oppositions might be discrete and mutually exclusive, such that one pole necessitates the complete lack of the other, such as dead versus alive. We might also have ensembles of multiple oppositions. For example, the Western meaning-system for direction consists of two binary oppositions of north/south and east/west, or of three oppositions (left/right, up/down, and front/back) (Geeraerts 2010:87). Oppositions may also have an evaluative component, such as good versus bad, clean versus dirty, and clever versus dull. In sum, the structuralist focus on *binary oppositions* (as opposed to other oppositions or other forms of meaning) may be overly reductionistic (Craib 1992).

*4.2.1. Do these critiques apply to word embeddings?*

Identifying binary oppositions – such as gender – in language data is a backbone of contemporary social science using word embeddings. Critiques of linguistic structuralism highlight that the focus on binary opposition in word embeddings may also be overly reductionistic. Methodological work highlighting the limits of measuring binary oppositions in word embeddings underscores this concern. Indeed, analysts have measured a wide variety of concepts as binary oppositions in semantic space, such as gender, age, and size. However, scholars find tremendous variation in the extent to which the resulting measures actually match human-rated perceptions of the concept (e.g., Chen, Peterson, and Griffiths 2017; Grand, G., Blank, I. A., Pereira, F., & Fedorenko, E 2018; Joseph and Morgan 2020). For example, when gender is measured as a binary opposition in embeddings, it appears to have an especially tight correspondence to human-ratings (Grand, G., Blank, I. A., Pereira, F., & Fedorenko, E 2018; Joseph and Morgan 2020), unlike, for example, race. Gender is a canonical case for studying concepts as binary oppositions in word embeddings. But perhaps gender is also an outlier in the extent to which manifests in our language as an opposition between two poles (see also Ethayarajh, Duvenaud, and Hirst 2019). While some concepts might be represented as a binary oppositional structure, other concepts might be better suited to some other structure.

Recently, some scholarship has begun to go beyond binary oppositions – looking for *systems* of oppositions and other latent structures in space. For instance, Boutyline et al. investigated gendered stereotypes relevant to education in print media from 1930-2009, including the gendered cultural associations of effort and intelligence (stereotypically feminine and masculine routes to success, respectively). Across time, the gendered associations of effort and intelligence became increasingly and synchronously polarized: as the former gained feminine associations the latter gain masculine associations (2020). Further, Kozlowski et al.

investigated class as a system of oppositions – investigating the relationships between five dimensions of class across time (2019). Even more recently, scholars have begun to investigate information structures beyond oppositions, such as looking for topical regions or clusters of words in semantic space (e.g., Arora et al. 2018; Arseniev-Koehler et al. 2020). Meanwhile, a small body of work investigates how to build word embeddings that can model more nuanced forms of information, such as hierarchy (Nickel and Kiela 2017). Thus, while early work using word embeddings heavily focused on binary oppositions, a stream of emerging work considers other structures.

*4.3. Critiques of Coherence*

The coherence posited by linguistic structuralism is also one of its most controversial aspects. Critics argue that structuralist perspectives envision meaning as unrealistically coherent (e.g., Bakhtin 1981; DiMaggio 1997; Martin 2010; Sewell 2005:169–72; Swidler 1986, 2013). Indeed, even among uses of a word within a given document, the word may evoke very different interpretations depending on its surrounding words. This phenomenon is known as polysemy. For instance, the word "depression" is entirely different in a sentence about mental health versus one about economics. The word "depression [economics]" is even more specific "The Great Depression," which refers to a particular economic depression. A word's meaning may also vary depending on a host of other extra-linguistic factors, such as where the text is produced and by whom, or who is reading the text (e.g., Franco et al. 2019; Geeraerts, Grondelaers, and Bakema 2012; Geeraerts and Speelman 2010; Hu et al. 2017; Peirsman, Geeraerts, and Speelman 2010; Robinson 2010). For instance, the word "awesome" evokes multiple distinct interpretations in different linguistic contexts; certain individuals are more likely to interpret "awesome" according to some of these interpretations than others, depending on their age, gender, and even

*DRAFT*     *Theoretical Foundations and Limits of Word Embeddings*

neighborhood (Robinson 2010). Such evidence for the variation of words highlights the shortcomings of linguistic structuralism, which focuses on language as a shared system.

*4.3.1. Do these critiques apply to word embeddings?*

Word embedding methods commonly used in social science, such as word2vec and GloVe, are vulnerable to the same longstanding critiques of coherence as linguistic structuralism. Most crucially, these models use a single word vector for each vocabulary word in the corpus, thus smoothing over the patterned ways in which a word, like depression, is used in the training corpus. Because these models are insensitive to linguistic context, they are commonly critiqued as modeling words as unrealistically coherent (e.g., Faruqui et al. 2016; Gladkova and Drozd 2016; Neelakantan et al. 2015; Wang et al. 2020). In fact, this limitations of models like word2vec and GloVe is well known, and prompted a new approach to model language in computer science: "contextualized" neural word embeddings (Devlin et al. 2019; M. Peters et al. 2018).

Contextualized embeddings models words in a way that presumes far less coherence. While models like word2vec and GloVe represent each *word* as a vector, contextualized models produce a vector for each *instance* of a word in a text. In a contextualized embedding, for example, each time the word "depression" is used in a corpus it may be modeled with slightly different vector; this vector is modified based on the context words used around each instance "depression" is mentioned. Thus, this approach enables words to vary across linguistic contexts.

To gain intuition into how a contextualized word embedding might address critiques of coherence, consider a simplified example of one approach[10] to contextualize word vectors. Like word2vec, an artificial neural network is tasked with "reading" a sentence and predicting the next word and initially represents each word as a static word vector. However, as the network predicts the next word in the sentence it *also* keeps an ongoing vector representing the "gist" of what is currently being talked about at any point in the sentence.[11] This "gist" is updated with each new word encountered in the sentence, and it is a function of the sequence of preceding words. Part of the model's training process is learning *how* to maintain this "gist": learning what information to keep, what to forget, and how to use information previously encountered, and as it reads a sentence and predicts a word. Once training is completed, we can input a sentence and, for any word used at some point *t* in the sentence, we can extract out the "gist" at time *t.* This "gist" is still represented as a vector, like our static word vectors, only now it is a function of the preceding context words in the sentence. By modifying the representation of the next word based on the current sentence, contextualized word embeddings address the critique that word embeddings like word2vec, model meaning as overly coherent.

For cultural sociologists, contextualized word embeddings do not merely offer a "solution" to analyze meaning in a way that responds to critiques of a coherent view of meaning. These approaches also enable us investigate *extent to which* meaning is coherent, aiding us to move past a dichotomous view of culture as either coherent or incoherent (Ghaziani 2011). Indeed, a recent study in computer science investigated the extent to which meaning is contextual in contextualized embeddings, by comparing words vectors from contextualized and non-contextualized embeddings (Ethayarajh 2019). This study found that, on average, less than 5% of meaning of a word's contextualized word vectors (where there is one word vector from each instance of the word in the corpus) could be explained by a single word vector. Further, contextualized models generally outperformed non-contextualized models on various linguistic tasks. At first glance, these results might suggest that enabling meaning to be incoherent more closely models human meaning, or that a coherent model of language is overly unrealistic. However, the extent to which contextualized embeddings outperform static models varies widely across specific linguistic tasks (Arora et al. 2020; Ethayarajh 2019; Tenney et al. 2019). Indeed, for many tasks and corpora, contextualized word embeddings only yields only marginal improvements (Arora et al. 2020; Tenney et al. 2019). In some of these cases, contextualized and non-contextualized embeddings even perform equally.[12] For social scientists, these findings suggest that the extent to which meaning is coherent is likely far more nuanced and remains an open (and promising) research area well suited to word embedding methods.

Still, even contextualized word embeddings do not fully overcome critiques of structuralist coherence. Like non-contextualized models, they remain insensitive to incoherence that may stem from other factors, such as who produces language, where and why. Like any word embedding, contextualized models learn to represent and process language from a training corpus and are thus tailored to the broader context of the training corpus. Therefore, while contextualized models offer a strategy to account for variation of meaning based on context words, neither contextualized nor non-contextualized word embeddings fully account for the many important ways in which meaning may be incoherent.

*4.4. Critiques of the Structuralist Focus on Language as Static*

The linguistic structuralist focus on language as static also received heavy critique. Linguistic structuralism distinguished the study of language across time from the study of language at a single time point, but then struggled to ever reconcile these two lenses (Giddens 1979:13; Stoltz 2019). Even if we give precedence to theorizing a symbolic system at a single time point, we also need to be able to explain changes in this system (Giddens 1979). As a theory, and even as a framework for empirical analysis, linguistic structuralism falls short at reconciling static and dynamic perspectives on language.

### 4.4.1. Do these critiques apply to word embeddings?

The extent to which word embeddings are vulnerable to critiques of static lens depends on the approach used to learn word embeddings: count-based versus artificial neural network based. Count-based word embeddings model a symbolic system as static, but require that the whole semantic space be abstracted at once by performing dimensionality reduction on a co-occurrence matrix, as described in section one. These methods do not incorporate any mechanism for change in a semantic space. Thus, count-based embeddings, like structuralist linguistics, fall short at reconciling a static and dynamic lens on language.

By contrast, *neural* word embeddings model language as a dynamic system that may be paused at any point for static analysis. Neural word embeddings are by-products of an online learning process. The word vectors are updated each time new cultural stimuli (e.g., text excerpts) are encountered, in a fashion that might be thought of as the process of socialization. Upon experiencing a context, the model uses its current information about each vocabulary word (i.e., "looks up" the word's position in the semantic space at this point) to make a prediction about the missing word in the context. Thus, word vectors structure how neural word embeddings experience any incoming language, are simultaneously structured by new experiences with language. When the prediction is incorrect, the positions of words are shifted, yielding an *updated* semantic space. In this way, neural word embeddings operationalize the notion a symbolic system, like any structure, is both a "thing" and a "process," i.e., a "structuring structure" (Giddens 1979; Sewell Jr 1992). The symbolic system captured by neural word embeddings is part of a dynamic process, and changes as the embedding interacts with its cultural environment.

When we use word vectors from neural word embeddings in social science applications, we generally stop this training process and begin analyses on the "frozen" system, as described earlier. We have extracted the word vectors as static representations from a system that can hypothetically change at any time with additional stimuli (i.e., additional text data) if we were to "unfreeze" the system. Thus, unlike count-based embeddings, neural word embeddings lend themselves to static analyses, but does not entirely divorce static and dynamic lenses.

Still, practical methods to study linguistic change using neural and count-based word embeddings are limited (Kutuzov et al. 2018). This limitation is important to address given the growing sociological interest in investigating culture across time using word embeddings (e.g., Boutyline et al. 2020; Jones et al. 2020; Kozlowski et al. 2019; Stoltz and Taylor 2020). Currently, one common approach to study linguistic change is to divide up the corpus into time segments and then compare embeddings trained on these separate segments (using count based or neural word embeddings) (e.g., Boutyline et al. 2020; W. L. Hamilton et al. 2016; Jones et al. 2020; Kozlowski et al. 2019; Stoltz and Taylor 2020). Then, to compare word vectors across time points, researchers may either (1) rotate embeddings from different segments so that their word vectors are directly comparable or (2) compare cosine similarities (i.e., between words or sets of words) in different segments. However, both approaches require large amounts of data to train on each segment. As a result, they may be unfeasible for many corpora sizes. Even on large corpora this approach does not allow for very granular time segments.

Another method to study linguistic change using embeddings is to train neural word embeddings on documents ordered across time: "freezing" and saving the system at various points in time, and then comparing the frozen models across this training (e.g., Kim et al. 2014). However, words' rate of semantic change might depend on their frequency in the corpus across time. Indeed, words which are not present for several time points might simply appear to have no semantic change. Further, the *quality* of word vectors may improve with increased training, making it challenging to disentangle training effects from true changes across time using this approach.

In sum, neural word embedding methods offer some solutions to model how a static semantic system may also change, but neither count-based nor neural word embeddings offer clear practical routes to empirically study language as a dynamic system. To overcome this limitation, additional methodological work on modeling linguistic change using word embeddings is still needed. Such methods will enable social scientists empirically analyze the structure and content of semantic systems across time in more precise ways.

## 5. DISCUSSION

Word embeddings open new doors for social scientists to investigate culture and language at scale. However, like any method, it is crucial that we clarify exactly what word embeddings operationalize. This paper has critically theorized how word embeddings may be used to operationalize several key premises of an influential theory of language: linguistic structuralism. Specifically, this paper focused on the linguistic structuralist premises that language is a relational and coherent symbolic system which is studied as if frozen in time.

This paper also theorized the ways in which word embeddings succumb to or overcome several critiques of these structuralist premises, such as debates about the (in)coherence of meaning and relational notions of meaning. As highlighted in this paper, different word embedding algorithms do so in different ways. In general, while count-based word embeddings have many of the same limitations of linguistic structuralism, neural word embeddings – and especially, contextualized neural word embeddings – offer solutions to these limitations. For example, neural word embeddings model a symbolic system as dynamic, which is then "frozen" for static, structuralist analyses. Further, while some word embeddings (e.g., word2vec) may model language as unrealistically coherent, contextualized neural word embeddings offer solutions to account for variation across linguistic contexts. More broadly, theoretical shortcomings of structuralism parallel advances in computer science to address the limitations of word embeddings. This includes the move from count-based to neural embeddings, the more recent move from static to contextualized embeddings, and the growing interests in diachronic and multimodal embeddings in computer science (see also Bisk et al. 2020).

The extent to which word embeddings succumb to critiques of structuralism also depends on the analysts' own interpretation of "meaning," in word embeddings. This includes, for example, whether the analyst interprets word embeddings as a *theoretical model*, or merely as a method to measure meaning. For instance, the distributional hypothesis may be seen as a modeling one mechanism by which meaning is constructed. Or the hypothesis may be interpreted as merely an approach to capture a proxy for meaning, "whatever meaning may be," thus side-stepping the concept of meaning altogether. Like network analysis (Borgatti et al. 2009) and other structuralist tools, word embeddings may be understood as a method or as a metaphor (Craib 1992:133). In sum, word embeddings *may* be used to operationalize structuralism, but it would be an oversimplification to say that all word embeddings, used across all analyses, are "structuralist."

*Directions for Future Social Science Research using Embeddings*

This paper hints at numerous future research directions that intersect computational social science and cultural sociology. First, embeddings are not only useful as to measure the *content* of meaning (e.g., gendered associations with career words). They are also useful to study the *structure* of meaning (e.g., morality as a binary opposition) and how the structure of meaning interacts with content and how this *structure* changes across time and space. For example, to what extent is morality represented in language as an opposition between good and evil, and to what extent is this opposition universal across languages? Or, perhaps morality actually described in language along multiple binary oppositions, such as sanctity/degradation and care/harm (Haidt and Joseph 2004; Sagi and Dehghani 2014). Importantly, we will need to distinguish our findings from the structure imposed by our measures. For example, measuring morality as a line implicitly limits the type of meaning we can find in semantic space: a word that is measured as more "good" with this method is necessarily less "evil." Thus, word embedding methods also require us to carefully consider both the content and structure of meaning.

Second, word embedding methods open new angles into longstanding debates about the coherence of meaning. Static and contextualized word embeddings represent two ends of this debate: while the former represents meanings as entirely static, the latter represent meaning as highly sensitive to its surrounding words. Contextualized word embeddings make it possible to compare meanings across linguistic contexts. These methods thus offer strategies to measure the extent of the variation of meaning, identify patterns the distribution of meanings (Sperber 1985), and perhaps ultimately explain the extent to which meaning is (in)coherent. At the same time, all word embeddings reflect their training corpus. Given that meaning may be incoherent across various possible training corpora, analysts must consider how the training corpus for *any* word embedding is produced, why, and by whom, and whose meanings the embedding represents – or excludes. In this way, longstanding debates about coherence in cultural sociology may be relevant to contemporary ethical issues in machine-learning, such as that language models often reflect the language (and ideologies about language) of dominant social groups (Blodgett et al. 2020; Shah, Schwartz, and Hovy 2020).

Third, extrapolating from the meaning of words to the meaning of other signs was a key and enduring legacy of linguistic structuralism, leading to the birth of the field of semiology more generally. Similarly, while word embeddings focus on the meaning in written language, their premises have been extended beyond modeling words, such as to model *nodes* in a social network (e.g., Grover and Leskovec 2016), and even *segments* in musical scores (Arronte Alvarez and Gómez-Martin 2019; e.g., Chuan et al. 2020). Many key theoretical points raised in this paper about word embeddings also extend to other modalities. For instance, all these variations of embeddings hinge on generalizing the Distributional hypothesis, where nodes, sounds, or images are defined by their relationship to other notes, sounds or images, respectively. This paper has focused on word embeddings given their recent rise in social science (e.g., Boutyline et al. 2020; Caliskan et al. 2017; Garg et al. 2018; Jones et al. 2020; Kozlowski et al. 2019). However, moving beyond word embeddings to other modalities could enable sociologists to consider how cultural signs, more generally, operate in our cultural environment (Bail 2014).

Finally, while this paper considers several core premises and critiques of linguistic structuralism (and structuralism more broadly) as they apply to word embeddings, this intellectual movement is broad (Dosse 1997). Future theoretical work might consider how word embeddings and their specific architectures align or diverge from these variations *within* linguistic structuralism, such as perspectives of de Saussure, Jakobson, and C.S. Pierce (e.g., Yakin and Totu 2014), and structuralism more broadly. Future work might also consider numerous critiques of these theories which are not covered in this paper, such as the role of agency and creativity. Such research might unveil other implicit theoretical assumptions – and potential innovations – in word embeddings.

*Conclusions*

Word embeddings are becoming pervasive social science approaches to analyze language, meaning, and culture using text data. However, these methods remain undertheorized. To ensure they are effectively used by social scientists, it is crucial that we define what *kind* of meaning word embeddings operationalize and their implicit assumptions. Dissecting the way that word embeddings implicitly formalize (or might be used to formalize) sociological concepts can ultimately push us to redefine these concepts themselves (Merton 1948). Analogously, social network analysis pushed scholars to clarify concepts like "social tie," "network," and "community" (Borgatti et al. 2009). Now, word embeddings offer a new theoretical opportunity to formalize concepts in cultural sociology, such as schema (Arseniev-Koehler and Foster 2020), binary opposition (Kozlowski et al. 2019), symbolic system, coherence, signs, and structure.

*DRAFT*                                    *Theoretical Foundations and Limits of Word Embeddings*

[1] It is standard practice when working with embeddings to normalize vectors in the space to have a length of 1.

[2] Measured with the implicit associations test (Greenwald, McGhee, and Schwartz 1998).

[3] Word2vec remains among the most popular and parsimonious neural word embedding models used in social science, and thus is one core neural word embedding model referenced in this paper. However, computer scientists have also developed a wide variety neural word embeddings which have specialized features; some of these variants are described as relevant in later sections (e.g., "contextualized" embeddings and multimodal word embeddings).

[4] Note that this explanation of word2vec is simplified for brevity and focuses on the version of word2vec which uses CBOW; for further details see Rong (2014) or Arseniev-Koehler (2020).

[5] In practice, this is not a exactly an average since high frequency words are downweighed in certain word2vec architectures (Arora et al. 2016).

[6] More precisely there are *two* word vectors for each word, one corresponding to contexts and one to target words. For details on the word2vec architecture, see Rong (2014)

[7] In structuralist jargon, these are referred to as "parole" versus "langue," respectively.

[8] In structuralist jargon, these are referred to as "synchronic" versus "diachronic" analyses.

[9] In practice, we usually stop the training process based on preset hyperparameters in the algorithm – such as error falling below a certain threshold or following a set cap on the number of iterations of the algorithm.

[10] The specific approaches to contextualize word vectors vary widely, but all use some form of an artificial neural network (for a review, see M. E. Peters et al. 2018).

[11] In more technical jargon, this conceptual description refers to the hidden state in a long-short-term memory (LSTM) network.

[12] Further, empirical work illustrates it is not necessarily the *contextualization* process that that leads to contextualized models' improvement on linguistic tasks (Arora et al. 2020) Indeed, contextualized embeddings also often include many other training architectures such as predicting sentences as well as words, accounting for suffixes and affixes, and accounting for the order of words. Compared to models like GloVe and word2vec, contextualized models also have enormous numbers of hyperparameters (i.e., knobs to tune to transform the inputted context words) and are trained on much larger sizes of corpora.